\documentclass[10pt,twocolumn,letterpaper]{article}

\usepackage{cvpr}
\usepackage{times}
\usepackage{epsfig}
\usepackage{graphicx}
\usepackage{epstopdf} % in order to convert eps files to pdf before running pdfLatex that otherwise will not recognize them
\usepackage{amsmath}
\usepackage{amssymb}
\usepackage{color}%for TODO

\usepackage[pagebackref=true,breaklinks=true,letterpaper=true,colorlinks,bookmarks=false]{hyperref}

 \cvprfinalcopy % *** Uncomment this line for the final submission

\def\cvprPaperID{134} % *** Enter the 3DV Paper ID here

\ifcvprfinal\fi
\begin{document}

\newcommand{\todo}[1]{\textcolor{red}{{\it TODO: #1}}}

% header with proceedings and info about the paper
\newcommand{\MYfooter}{\smash{\scriptsize
\hfil\parbox[t][\height][t]{\textwidth}{\centering
Accepted version of paper published at 3DV 2014, http://dx.doi.org/10.1109/3DV.2014.46}\hfil\hbox{}}}

\makeatletter

% normal pages
\def\ps@headings{%
\def\@oddhead{\mbox{}\scriptsize\rightmark \hfil \thepage}
\def\@evenhead{\scriptsize\thepage \hfil \leftmark\mbox{}}
\def\@oddfoot{\MYfooter}%
\def\@evenfoot{\MYfooter}}

% title page
\def\ps@IEEEtitlepagestyle{%
\def\@oddhead{\mbox{}\scriptsize\rightmark \hfil \thepage}%
\def\@evenhead{\scriptsize\thepage \hfil \leftmark\mbox{}}%
\def\@oddfoot{\MYfooter}%
\def\@evenfoot{\MYfooter}}

\makeatother

% make changes take effect
\pagestyle{headings}
% adjust as needed
\addtolength{\footskip}{0\baselineskip}
\addtolength{\textheight}{-1\baselineskip}

% GENERAL COMMANDS-------------------------------------------------------------
%additional material {\tt fg324.pdf}
%something~\cite{Alpher02,Alpher03,Authors12}.
%Alpher \etal~\cite{Alpher04}
%$\mathit{conf}_a$
%bla\footnote{footnote}
%\eg,

% FIGURES COMMANDS--------------------------------------------------------------
%   \usepackage[dvips]{graphicx} ...
%  \includegraphics[width=0.8\linewidth]
%                  {myfile.eps}

%\begin{figure*}
%\begin{center}
%\fbox{\rule{0pt}{2in} \rule{.9\linewidth}{0pt}}
%\end{center}
%   \caption{Example of a short caption, which should be centered.}
%\label{fig:short}
%\end{figure*}

%\begin{figure}[t]
%\begin{center}
   %\includegraphics[width=0.8\linewidth]{egfigure.eps}
%\end{center}
%  \caption{Formulas ($B \sin A = A \sin B$).}
%\label{fig:long}
%\label{fig:onecol}
%\end{figure}

%\begin{table}
%\begin{center}
%\begin{tabular}{|l|c|}
%\hline
%Method & Frobnability \\
%\hline\hline
%Theirs & Frumpy \\
%Yours & Frobbly \\
%Ours & Makes one's heart Frob\\
%\hline
%\end{tabular}
%\end{center}
%\caption{Results.   Ours is better.}
%\end{table}

%%%%%%%%% TITLE%%%%%%%%%%%%%%%%%
\title{Efficient Multi-view Performance Capture of Fine-Scale Surface Detail}

\author{Nadia Robertini\textsuperscript{1,2}, Edilson De Aguiar\textsuperscript{1,3}, Thomas Helten\textsuperscript{1,4}, Christian Theobalt\textsuperscript{1}\\
{\small \textsuperscript{1}MPI Informatik, Germany\hspace{10pt}\textsuperscript{2}Intel VCI, Germany\hspace{10pt}\textsuperscript{3}CEUNES / UFES, Brazil\hspace{10pt}\textsuperscript{4}Pixargus / GoalControl, Germany}\\
{\tt\small \{nroberti,edeaguia,thelten,theobalt\}@mpi-inf.mpg.de}\\
%{\small\url{http://www.author.org/~second}}
%\and
%Edilson De Aguiar\\
%MPI Informatik\\
%\and
%Thomas Helten\\
%MPI Informatik\\
%\and
%Christian Theobalt\\
%MPI Informatik\\
}

\maketitle
% \thispagestyle{empty}

%%%%%%%%% ABSTRACT
\begin{abstract}

We present a new effective way for performance capture of deforming meshes with fine-scale time-varying surface detail 
from multi-view video. Our method builds up on coarse 4D surface reconstructions, as obtained with commonly used template-based methods. 
As they only capture models of coarse-to-medium scale detail, fine scale deformation detail is often done in a second pass by using stereo constraints, features, or shading-based refinement. 
 In this paper, we propose a new effective and stable solution to this second step. 
%We phrase dense dynamic surface reconstruction 
%as a global optimization problem of the densely deforming surface.  
%Starting with a synchronized multiple video recording of a subject performing a motion and its reconstructed coarse 
%mesh animation, our technique reconstructs dense fine-scale surface details on the input mesh animation according to 
%photo-consistency constraints. 
Our framework creates an implicit representation of the deformable mesh using a dense 
collection of 3D Gaussian functions on the surface, and a set of 2D Gaussians for the images. 
The fine scale deformation of all mesh vertices that maximizes photo-consistency can be efficiently found by densely optimizing a new model-to-image consistency energy  
on all vertex positions. 
A principal advantage is that our problem formulation yields a smooth closed form 
energy with implicit occlusion handling and analytic derivatives. Error-prone correspondence finding, 
or discrete sampling of surface displacement values are also not needed. We show several reconstructions of human subjects wearing loose 
clothing, and we qualitatively and quantitatively show that we robustly capture more detail than related methods. 

\textbf{Acknowledgement.} This project was supported by the ERC Starting Grant 335545 CapReal and CAPES Grant 11012-13-7.
\end{abstract}

%%%%%%%%% BODY TEXT
\section{Introduction}
\begin{figure}
\begin{center}
 \includegraphics[width=0.8\linewidth]{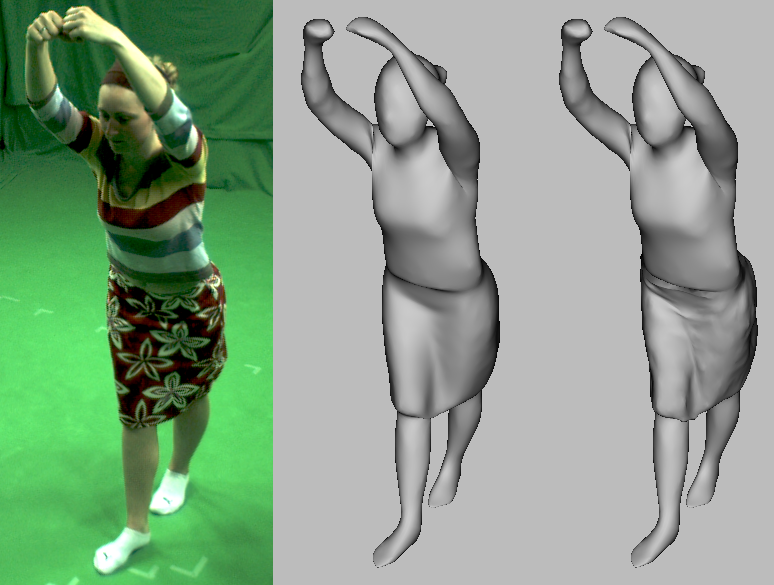}
\end{center}
   \caption{Given as input a multi-view video sequence (left - particular input frame) and a coarse mesh animation (middle - input mesh), our method is able to efficiently reconstruct 
   fine scale surface details (right - refined mesh). Note the wrinkles and folds reconstructed on the skirt.}
\label{figure:teaser}
\end{figure}
Performance capture methods enable the reconstruction of the motion, the dynamic surface geometry, and the appearance of real 
world scenes from multiple video recordings, for example, the deforming geometry of body and apparel of an actor, 
or his facial expressions~\cite{Aguiar2008, Gall2009, Bradley2008, Vlasic08}. 
Many methods to capture space-time coherent surfaces reconstruct a coarse-to-medium scale 4D model of the 
scene in a first step, \eg by deforming a mesh or a rigged template such that it aligns with the images~\cite{Aguiar2008,Vlasic08}.
Finer scale shape detail is then added in a second refinement step. In this second step, some methods align the surface to a combination of silhouette constraints 
and sparse image features~\cite{Gall2009}. But such approaches merely recover medium scale detail and may suffer from erroneous feature correspondences between images and shape. 
Photo-consistency constraints can also be used to compute smaller scale deformations via stereo-based refinement~\cite{Aguiar2008,StarckHilton2007}.
However, existing approaches that follow that path often resort to discrete sampling of local displacements, since phrasing dense stereo based refinement as a continuous optimization problem has been 
more challenging~\cite{Kolev-et-al-ijcv09}.    
Some recent methods resort to shading-based techniques to capture small-scale displacements, such as shape-from-shading or photometric stereo~\cite{Wu2011,Wu2011TVCG,Vlasic08}.
%\todo{%also cite Vlasic/Debevec here, 
%maybe even the Hernandez paper}. 
However, the methods either require controlled and calibrated lighting, or complex inverse estimation of lighting and appearance when they are applied under 
uncontrolled recording conditions. 

In this paper, we contribute with a new effective solution to the refinement step using multi-view photo-consistency constraints. 
As input, our method expects synchronized and calibrated multiple video of a scene and a reconstructed coarse mesh animation, as it can be obtained 
with previous methods from the literature. Background subtraction or image silhouettes are not required for refinement. 

Our first contribution is a new shape representation that models the mesh surface with a dense collection of 3D Gaussian functions centered at each vertex 
and each having an associated color. A similar decomposition into 2D Gaussian functions is applied to each input video frame. 

This scene representation enables our second contribution, namely the formulation of dense photo-consistency-based surface refinement as a global optimization 
problem in the position of each vertex on the surface. Unlike previous performance capture methods, we are able to phrase the model-to-image photo-consistency energy that guides the deformation 
as a closed form expression, and we can compute its analytic derivatives. Our problem formulation has the additional advantage that it enables implicit handling of occlusions, as well as spatial and temporal coherence constraints, while preserving a smooth consistency energy function. We can effectively minimize this function in 
terms of dense local surface displacements with standard gradient-based solvers. In addition to these advantages, unlike many previous methods, our framework does not require a potentially error-prone sparse set of feature correspondences or discrete sampling and testing of surface displacements, and thus provides a new way of continuous optimization of the dense surface deformation.

We used our approach for reconstructing full-body performances of human actors wearing loose clothing, and performing different motions. 
Initial coarse reconstructions of the scene were obtained with the approaches by Gall \etal~\cite{Gall2009} and Starck and Hilton~\cite{StarckHilton2007}.
Our results (Fig.~\ref{figure:teaser} 
and Sect.~\ref{section:results}) show that our approach is able to reconstruct more of the fine-scale detail that is present in the input video sequences, than the 
baseline methods, for instance the wrinkles in a skirt. We also demonstrate these improvements quantitatively. 

%-------------------------------------------------------------------------
\section{Related Work}
%Marker-less motion capture approaches are able to reconstruct human skeletal motion without attaching markers to the 
%body of an actor~\cite{MOHK06, POPP07, HEVA10}. 
%Most approaches are based on a template skeleton with simple attached 
%shape primitives and therefore are not able to capture the deforming surface along with the 
%skeleton\cite{DEBR00,BRMP04,GARS07,HEVA10,STOLL11}. By extending marker-less motion capture algorithms fro%m computer 
%vision, the simultaneously capture of the motion, appearance and shape details is accomplished by performance capture methods. 

Marker-less performance capture methods are able to reconstruct dense dynamic surface geometry of moving subjects from multi-view video, for instance of people in loose clothing, possibly along with pose parameters of 
an underlying kinematic skeleton~\cite{Theobalt_Springer_2010}. Most of them use data from dense multi-camera systems and recorded under controlled studio environments. Some methods employ variants of shape-from-silhouette or active or passive stereo~\cite{Zitnick:Sig04,IBVH,StarckHilton2007,Was05,Tung09}, which usually results in temporally incoherent reconstructions.  
%Wilson et al~\cite{wilson:acmcg2010} use stereo and optical flow in a light-stage setup to obtain a temporally coherent parameterization for facial performance capture. They compute optical flow amidst a subset of {\em tracking frames} that are all captured under the same incident lighting. 
Space-time coherency is easier to achieve with model-based approaches that deform a static shape template (obtained by a laser scan or image-based reconstruction) such that it matches the subject, \eg a person~\cite{CarrTheoSig03,Aguiar2008,Vlasic08,Ballan3DPVT08,Gall2009} or a person's apparel~\cite{Bradley2008}. Some of them jointly track a skeleton and the non-rigidly deforming surface~\cite{Vlasic08,Ballan3DPVT08,Juergen09}; also multi-person reconstruction has been demonstrated~\cite{LiuTPAMI2013}.
Other approaches use a generally deformable template without embedded skeleton to capture 4D models, \eg an elastically deformable surface or volume~\cite{Aguiar2008,Savoye:2013}, or a patch-based surface representation~\cite{CagniartCVPR10}. Most of the approaches mentioned so far either only reconstruct coarse dynamic 
surface models that lack fine scale detail, or coarse reconstruction is a first stage. Fine scale detail is then added to the coarse result in a second refinement step. 

Some methods use a combination of silhouette constraints and sparse feature correspondences to estimate, at best, 
a medium scale non-rigid 4D surface detail~\cite{Gall2009}. Other approaches use stereo-based photo-consistency constraints in addition to silhouettes to achieve denser estimates of finer scale deformations~\cite{StarckHilton2007,Aguiar2008}. It is an involved problem to phrase dense stereo-based surface refinement as a continuous optimization problem, as it is done in variational approaches~\cite{Kolev-et-al-ijcv09}. Thus, stereo-based refinement in performance capture often resorts to discrete surface displacement sampling which are less efficient, and with which globally smooth and coherent solutions are harder to achieve.

In this paper, we propose a new formulation of stereo-based surface refinement as a continuous optimization problem, which is based on a new surface representation with Gaussian functions. In addition, our refinement method also succeeds if silhouettes are not available, making the approach more generally applicable.  

An alternative way to recover fine-scale deforming surface detail is to use shading-based methods, \eg shape-from-shading 
or photometric stereo~\cite{Wu2011TVCG}. Many of these approaches require controlled and calibrated lighting~\cite{Hernandez07,VlasicSIGASIA09}, which reduces their applicability. More recently, shading-based refinement of dynamic scenes captured under more general lighting was shown~\cite{Wu2011}, but these approaches are computationally challenging as they require to solve an inverse rendering problem to obtain estimates of 
illumination, appearance and shape at the same time.

The method we propose has some similarity to the work of Sand \etal ~\cite{Sand:2003} who capture skin deformation as a displacement field on a template mesh; however, they require marker-based skeleton capture, and only fit the surface to match the silhouettes in  multi-view video.  
Our problem formulation is inspired by the work of Stoll \etal~\cite{Stoll2011} who used a collection of Gaussian functions 
in 3D and 2D for marker-less skeletal pose estimation. Estimation of surface detail was not the goal of that work. Our paper extends their basic concept to the different problem of dense stereo-based surface estimation using continuous optimization of a smooth energy that can be formulated in closed form, and that has analytic derivatives. 
%In~\cite{Stoll2011}, they propose an approach for the reconstruction of the human body motion and deforming surface. 
%They employ a Sums-of-Gaussians body model, where its low-frequency surface is defined implicitly through a set of colored 
%Gaussians functions, or blobs, and solve an energy that aims to maximize color consistency between the model and the images. 
%However, while they focus on the coarse reconstruction of the human pose, we extended the original framework to reconstruct dense 
%surface details.

%-------------------------------------------------------------------------
\section{Overview}
\label{sect:overview}
\begin{figure*}[t!]
\begin{center}
\includegraphics[width=0.8\linewidth]{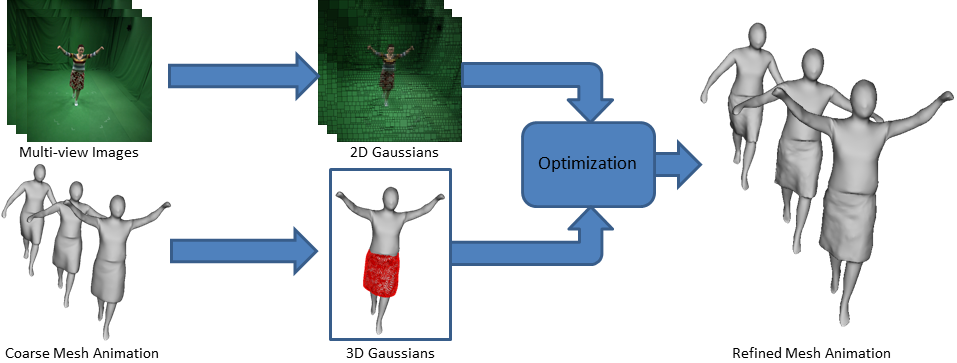}
\end{center}
 \caption{Overview of our framework. Our approach refines the input coarse mesh animation by maximizing the color consistency between the 
 collection of 3D surface Gaussians, associated to the input vertices, and the set of 2D image Gaussians, assigned to image patches.}
\label{figure:overview}
\end{figure*}

An overview of our approach is shown in Fig.~\ref{figure:overview}. The input to our algorithm is a calibrated and synchronized 
multi-view video sequence showing images of the human subject.  In addition, 
we assume as input a spatio-temporally coherent coarse animated mesh sequence, reconstructed from multi-view video
related approaches~\cite{Gall2009,StarckHilton2007}.

Our method refines the initial coarse animation such that the fine dynamic surface details are incorporated to the meshes.  
First, we create an implicit representation of the input mesh using a dense collection of 3D Gaussian functions on the surface with 
associated colors. 
The input images are also represented as a set of 2D Gaussian associated to image patches in each camera view. 
Thereafter, continuous optimization is performed to maximize the color consistency between the collection of 3D surface Gaussians and 
the set of 2D image Gaussians.  The optimization displaces the 3D Gaussians along the associated vertex normal of the coarse mesh 
which yields the necessary vertex displacement.

Our optimization scheme has a smooth energy function, that, thanks to our Gaussians-based model, can be expressed
in closed form. It further allows us to analytically compute derivatives, enabling the possibility of using efficient gradient-based
solvers.

%-------------------------------------------------------------------------
\section{Implicit Model}
Our framework converts the input coarse animation and input multi-view images into implicit representations using a collection 
of Gaussians: 3D surface Gaussians on the mesh surface with associated colors and 2D image Gaussians, with associated colors, assigned to image patches in each camera view.

\subsection{3D Surface Gaussian}
\begin{figure}[t]
\begin{center}
  \includegraphics[width=0.6\linewidth]{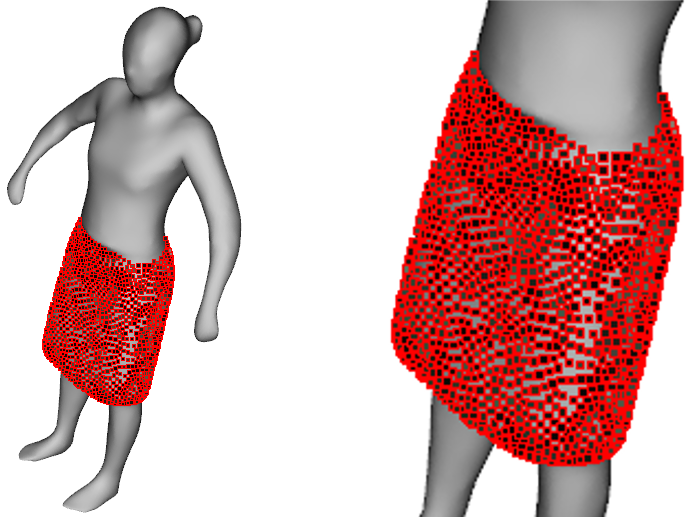}
\end{center}
\caption{A representation of our collection of 3D Gaussian on the surface. Our surface Gaussians, here illustrated as tiny red-bordered squares,
are assigned to vertices of the input coarse mesh (only in the skirt region in this example). %Their color attribute is computed by averaging the underlying pixel colors of the correspondent Gaussian projection in the best camera view.
%and associated to the projected vertex color in the best camera view.
}
\label{figure:modelGaussians}
\end{figure}

Our implicit model for the input mesh is obtained by placing a 3D Gaussian at each mesh vertex $v_s$, $\forall s \in \{0\dots n_s-1\}$, $n_s$ being the number of vertices.
A 3D un-normalized isotropic Gaussian function on the surface is defined simply with a mean \(\hat{\mu}_s\), that coincides with the vertex location,
and a standard deviation \(\hat{\sigma}_s\) (equally set to \(7\) \(mm\) for all 3D Gaussians on surface) as follows:
\begin{equation}
G_s(\hat{x}) = exp\left(-\frac{||\hat{x}-\hat{\mu}_s||^2}{2 \hat{\sigma}_s^2}\right)
\end{equation}
\indent with \(\hat{x} \in \mathbb{R}^3\).
Note that although \(G_s(\hat{x})\) has infinite support, for visualization purposes we represent its projection as a square having center (\ie diagonals intersection) in \(\hat{\mu}_s\) and side length equal to \(2 \hat{\sigma}_s\) \(mm\) (see Fig.~\ref{figure:modelGaussians}).

We further assign a HSV color value \(\eta_s\) to each surface Gaussian.
In order to derive the colors we choose a reference frame where the initial coarse reconstruction is as close as possible to the real shape. 
This is typically the first frame in each sequence.
For each vertex $v_s$ of the input mesh, we first choose the camera view that sees vertex $v_s$ best, \ie where normal and camera viewing direction align best.
Thereafter, the 3D Gaussian associated to $v_s$ is projected to the image from the best camera view and the underlying pixel color average is assigned as a color attribute.

%-------------------------------------------------------------------------
\subsection{2D Image Gaussian}
\label{section:2dimagegaussian}
\begin{figure}[t]
\begin{center}
\includegraphics[width=0.45\linewidth]{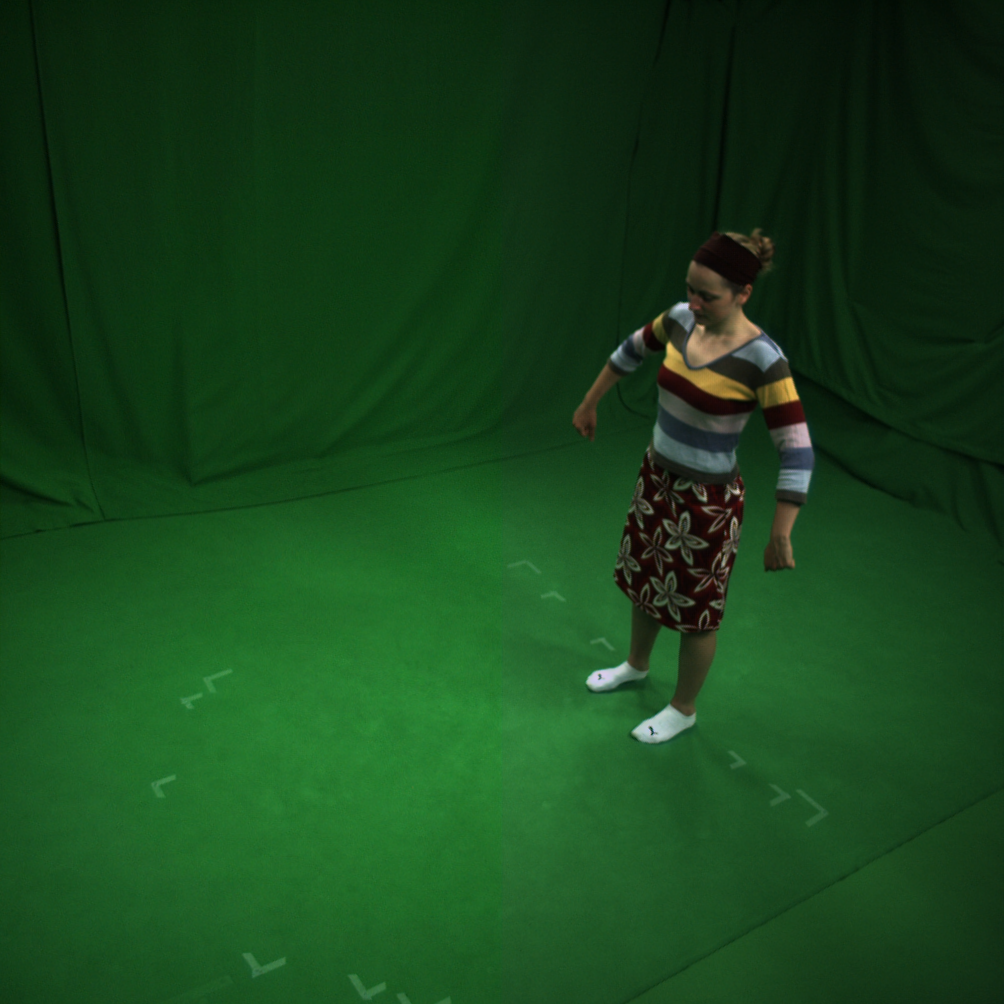}
\includegraphics[width=0.45\linewidth]{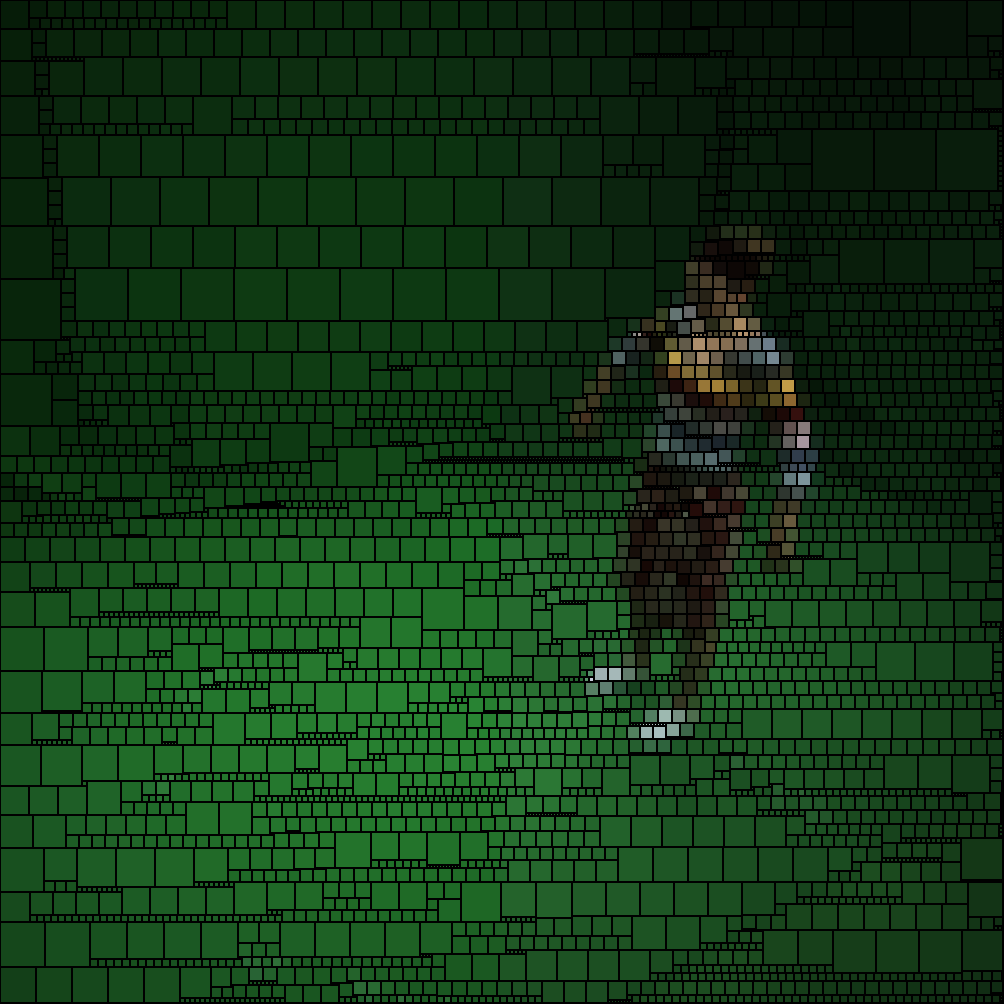}
\end{center}
\caption{The input image (left) and the estimated collection of 2D image Gaussians (right). The image Gaussians are assigned to patches of coherent color in the input image and the
underlying average pixel color is assigned to the Gaussians as additional attribute.
%Our 2D Image Gaussians are assigned to patches in the images., represented as clusters with the assigned underlying pixel color.
}
\label{figure:imageGaussians}
\end{figure}

Our implicit model for the input images of all cameras $c \in \{0\dots n_c-1\}$, $n_c$ being the number of cameras, is obtained by assigning 2D Gaussian functions $G_i(x)$, $x \in \mathbb{R}^2$, to each image patch, $i \in I(c)$, of all camera views. %=\{0\dots n_i^c-1\}$.
Similar to Stoll \etal~\cite{Stoll2011} we decompose each input frame into squared regions of coherent color by means of quad-tree decomposition (with maximal depth set to 8).
A 2D Gaussian is assigned to each patch (Fig.~\ref{figure:imageGaussians}), such that its mean \(\mu_i \in \mathbb{R}^2\) corresponds to the patch center, and its standard deviation
\(\sigma_i\) to half of the square patch side length. The underlying average HSV color \(\eta_i\) is also assigned to the 2D Gaussians as additional attribute.

%-------------------------------------------------------------------------
\subsection{Projection of 3D Surface Gaussians}
In order to evaluate the similarity between the 3D surface Gaussians $G_s$ and the 2D image Gaussians $G_i$, we project each $G_s$ to the 2D image space. 
The 3D surface Gaussian mean \(\hat{\mu}_s\) is projected using the camera projection matrix \(P\), similarly to any 3D point in the space, as follows:
\begin{equation}
\mu_{s} = 
\left(
    \begin{array}{c}
     \frac{[P \hat{\mu}_s^h]_x}{[P \hat{\mu}_s^h]_z}  \\
\\
     \frac{[P \hat{\mu}_s^h]_y}{[P \hat{\mu}_s^h]_z} 
    \end{array}
\right)  \in \mathbb{R}^2
\end{equation}
with $[P \hat{\mu}_s^h]_{x,y,z}$ being the respective coordinates of the projected mean in homogeneous coordinates (\ie the $4^{th}$ dimension is set to $1$).
%We approximate the projected 3D Gaussian as an uniform 2D Gaussian, and project the 3D standard deviation as follows:
The 3D standard deviation is projected using the following formula:
\begin{equation}
\sigma_s = \frac{\hat{\sigma}_s f}{[P \hat{\mu}_s^h]_z} \in \mathbb{R}
\end{equation}
where \(f\) is the camera focal length.

%-------------------------------------------------------------------------
\section{Surface Refinement}
\label{sect:refinement}
We employ an analysis-by-synthesis approach to refine the input coarse mesh animation, at every frame, by optimizing the following energy \(\textbf{E}(M)\) 
with respect to the collection of 3D surface Gaussian means \(M = \{\hat{\mu}_0,\dots \hat{\mu}_{n_s-1}\}\):
\begin{equation}
\textbf{E}(M) = E_{sim} - w_{reg} E_{reg} 
\label{eq:energy}
\end{equation}

%Our energy function has two terms: 
The term \(E_{sim}\) measures the color similarity of the projected collection of 3D surface Gaussians with the 
2D image Gaussians obtained from each camera view.  The additional term \(E_{reg}\) is used to keep the distribution of the 3D surface Gaussians geometrically smooth,
whereas \(w_{reg}\) is an user defined smoothness weight, typically set to 1.
Since we constrain the 3D Gaussians to move along the corresponding vertex (normalized) normal direction \(N_s\):
\begin{equation}
\label{eq:mu}
\hat{\mu}_{s} = \hat{\mu}_{s}^{orig} + N_s k_s \in \mathbb{R}^3
\end{equation}
aiming at maintaining a regular distribution of 3D Gaussians on the surface, we only need to optimize for single scalar values \(k_s, s \in \{0\dots n_s-1\}\).

%-------------------------------------------------------------------------
\subsection{Similarity Term}
We exploit the power of the implicit Gaussian representation of both input images and surface in order to derive a closed-form analytical
formulation for our similarity term.
%A good formulation for evaluating the similarity of the image Gaussians and the 3D Gaussians projected into all camera views,
%can be derived by exploiting the power of the implicit Gaussian functions. 
In principle, one pair of image Gaussian and projected surface Gaussian
should have high similarity measures when they show similar properties in terms of color
and their spacial localization is sufficiently close. This measure can be formulated as the integral of the product of the projected 
surface Gaussian $G_s(x)$ and image Gaussian $G_i(x)$, weighted by their color similarity \( T(\delta_{is})\), as follows:
\begin{equation}
%\begin{split}
E_{is} = T(\delta_{is}) \int_{\Omega}{G_i(x)  G_s(x) \partial{x}}
%&= d(c_i,c_s) 2 \pi \frac{\sigma_{s}^2\sigma_{i}^2}{\sigma_{s}^2+\sigma_{i}^2} \hspace{5px} exp\left(-\frac{||\mu_i - \mu_s ||^2}{\sigma_{s}^2+\sigma_{i}^2}\right)
%\end{split}
\label{equation:gaussianOverlap}
\end{equation}
In the above equation $\delta_{is} = || \eta_i - \eta_s || \in \mathbb{R}^{+}$ measures the Euclidean distance between the colors, while \(T(\delta): \mathbb{R} \to \mathbb{R}\) is the Wendland radial basis function modeled by:
\begin{equation}
T(\delta) = \left\{
\begin{array}{l l}
\Big(1 - \frac{\delta}{\Delta}\Big)^4 \Big(4 \frac{\delta}{\Delta} + 1\Big) & \quad \text{if $\delta < \Delta$}\\
&\\ % leaves some space
0 & \quad \text{otherwise}
\end{array} \right.
\label{equation:T}
\end{equation}
where $\Delta$ is esperimentally set to $0.05$ for all test sequences.
The main advantage of using a Gaussian representation is that the integral in Eq.~\ref{equation:gaussianOverlap} has a closed-form solution, namely another Gaussian with combined properties:
\begin{equation}
%\begin{split}
E_{is} = T(\delta_{is}) 2 \pi \frac{\sigma_{s}^2\sigma_{i}^2}{\sigma_{s}^2+\sigma_{i}^2} \hspace{5px} exp\left(-\frac{||\mu_i - \mu_s ||^2}{\sigma_{s}^2+\sigma_{i}^2}\right)
%\end{split}
\label{equation:gaussianOverlapExplicit}
\end{equation}

We first calculate the similarity for all components of the two models for each camera view. Then, we normalize the result considering the maximum obtainable overlap \(E_{ii} = \pi \sigma_{i}^2\), of an image Gaussian with itself, and the number of cameras \(n_c\) as follows:
\begin{equation}
%E_{sim} = \frac{1}{n_{c}} \sum_{c=0}^{n_{c}-1} \frac{1}{\sum_{i\in I(c)} E_{ii}} \sum_{i \in I(c)} min \left(\sum_{s = 0}^{n_s-1} E_{is},E_{ii}\right)
E_{sim} = \frac{1}{n_{c}} \sum_{c=0}^{n_{c}-1} \sum_{i \in I(c)} \frac{min \left(\sum_{s = 0}^{n_s-1} E_{is},E_{ii}\right)}{E_{ii}}
\label{equation:similarity}
\end{equation}

In this equation, the inner minimization implicitly handles occlusions on the surface as it prevents occluded Gaussians projections into the same image location
to contribute multiple times to the energy. This is an elegant way for handling occlusion while preserving at the same time energy smoothness. In fact, exact occlusion detection and handling algorithms are non-smooth or hard to express in closed-form.

In order to improve computational efficiency, we evaluate $E_{is}$ only for visible surface Gaussians from each camera view.
The Gaussian overlap is then computed against visible projected Gaussians and 2D image Gaussians in a local neighborhood.
%We also only evaluate \(E_{is}\) against a projected Gaussian and 2D image Gaussians in a local neighborhood.
%-------------------------------------------------------------------------
\subsection{Regularization Term}
Our regularization term constraints the 3D surface Gaussians in the local neighborhood and each Gaussian such that the final reconstructed surface is sufficiently smooth. 
This is accomplished by minimizing the following equation:
\begin{equation} \label{eq:reg}
E_{reg} = \sum_{s = 0}^{n_s-1} {\sum_{j \in \Psi(s)} T(\delta_{sj}) \left(k_s - k_j\right)^2},
\end{equation}

where \(\Psi(s)\) is a set of surface Gaussian indices that are neighbors of \(G_s\), $\delta_{sj} \in \mathbb{R}^{+}$ is the geodesic surface distance between \( G_s \) and \( G_j \)
measured in number of edges, and \(T(\delta)\) is defined in Eq.~\ref{equation:T}, where $\Delta = 2\hspace{0.1cm}edges$.
%In Figure \ref{figure:regularization} we show the effects of tweaking of \(M_d\) and \(w_{reg}\).
%\begin{figure}[t]
%\begin{center}
%\includegraphics[width=0.22\linewidth]{images/Regularization0.png}
%\includegraphics[width=0.22\linewidth]{images/Regularization1.png}
%\includegraphics[width=0.22\linewidth]{images/Regularization2.png}
%\includegraphics[width=0.22\linewidth]{images/Regularization3.png}
%\end{center}
%\caption{Effect of the refinement term in the reconstruction result.  From left to right: (1) \(M_d = 5\) and \(w_{reg} = 1\), (2) \(M_d = 2\) and \(w_{reg} = 1\), (3) \(M_d = 1\) and \(w_{reg} = 1\), (4) \(w_{reg} = 0\). We chose configuration settings (2) for all our test sequences, as they give best compromise for accuracy and surface smoothness.
%}
%\label{figure:regularization}
%\end{figure}

%-------------------------------------------------------------------------
\subsection{Optimization}
%\todo{Gaussian functions are a good alternative since they are infinitely smooth and, therefore, have smooth derivatives. 
%Smooth derivatives is a great advantage when we need to maximize an energy function, as the solution can be approximated 
%numerically and very efficiently. Another advantage is that the overlap with other Gaussian function is easy to compute, 
%having an elegant formulation.}
%Our energy can be efficiency optimized using gradient based approaches, \eg gradient ascent. We use a conditioned gradient approach
%aiming at faster convergence.
Our formulation allows us to compute analytic derivatives of the energy (Eq.~\ref{eq:energy}), for which we provide complete
derivation in an additional document. The derivative of the similarity term, with respect to each $k_s, s \in \{0\dots n_s - 1\}$ is:
\begin{equation}
\begin{split}
%\frac{\partial}{\partial k_s}(E_{sim}) = \frac{1}{n_{c}} \sum_{c=0}^{n_{c}-1} \frac{1}{\sum_{i=0}^{n_i} E_{ii}} \sum_{i = 0}^{n_i} \Phi(k_s) % \left\{
\frac{\partial}{\partial k_s}&(E_{sim}) =\\
&\frac{1}{n_{c}} \sum_{c=0}^{n_{c}-1} \sum_{i\in I(c)} 
\left\{
\begin{array}{l l}
\frac{\frac{\partial}{\partial k_s}(E_{is})}{E_{ii}} & \text{if} \sum_{s = 0}^{n_s-1} E_{is} < E_{ii}\\
&\\ % leaves some space
0 & \text{otherwise}
\end{array} \right.
\end{split}
\end{equation}

The derivative of the overlap \(E_{is}\) is defined as:
\newcommand{\MuX}{{[\mu_{s}]}_x} % X COMPONENT
\newcommand{\MuY}{{[\mu_{s}]}_y} % Y COMPONENT
\newcommand{\MuZ}{{[\mu_{s}]}_z} % Z COMPONENT
\newcommand{\SumVar}{\sigma_{s}^2+\sigma_{i}^2} % sum of variance square
\newcommand{\DistMean}{||\mu_i - \mu_s ||^2} % dist mean square
\newcommand{\ProdExp}{\frac{\sigma_{s}^2\sigma_{i}^2}{\sigma_{s}^2+\sigma_{i}^2} exp\left(-\frac{||\mu_i - \mu_s ||^2}{\sigma_{s}^2+\sigma_{i}^2}\right)} % product of division and exponential
\newcommand{\DProdExp}{2 \ProdExp \Bigg[\frac{\partial}{\partial k_s}(\MuZ) \Bigg(-1 +\\
&+ \frac{\sigma_{s}^2}{\SumVar} - \frac{\DistMean \sigma_{s}^2}{(\SumVar)^2 }\Bigg)\frac{1}{\MuZ} + \frac{(\mu_i - \mu_s) \frac{\partial}{\partial k_s}(\mu_{s})}{\SumVar}\Bigg]}  %derivative of product of division and exponential
\begin{equation}
\label{eq:final}
\begin{split}
\frac{\partial}{\partial k_s}&(E_{is}) = T(\delta_{is}) 4 \pi \ProdExp \cdot \\
& \cdot \Bigg[[P_c N_s^h]_z \Bigg(-1 + \frac{\sigma_{s}^2}{\SumVar} - \frac{\DistMean \sigma_{s}^2}{(\SumVar)^2 }\Bigg) \cdot\\
&\cdot \frac{1}{\MuZ} + \frac{(\mu_i - \mu_s) \frac{\partial}{\partial k_s}(\mu_s)}{\SumVar}\Bigg]
\end{split}
\end{equation}

where $P_c$ is the projection matrix of camera $c$, $N_s^h$ is the vertex normal associated to the model gaussian $G_s$ in homogeneous coordinates (\ie the $4^{th}$ dimension is set to $0$), $[\mu_s]_z$ is the z-component of the projected mean, and% $\frac{\partial}{\partial k_s}(\mu_{s})$ is:
\newcommand{\MuN}{\hat{\mu}_s^h + N_s^h k_s} % mu along its vertex normal
\newcommand{\DMuX}{{[P N_s^h]}_x} % X COMPONENT
\newcommand{\DMuY}{{[P N_s^h]}_y} % Y COMPONENT
\newcommand{\DMuZ}{{[P N_s^h]}_z} % Z COMPONENT
\begin{equation}
\frac{\partial}{\partial k_s}(\mu_{s}) = 
\left(
    \begin{array}{c}
     {\frac{\DMuX - [\mu_s]_x \DMuZ}{[P (\MuN)]_z}}\\
\\
     {\frac{\DMuY - [\mu_s]_y \DMuZ}{[P (\MuN)]_z}}\\
    \end{array}
\right) \in \mathbb{R}^2.
\end{equation}
The derivative of the regularization term \(E_{reg}\) is given by:
\begin{equation}
\begin{split}
\frac{\partial}{\partial k_s}&(E_{reg}) = 4 \sum_{j \in \Psi(s)} T(\delta_{sj}) \left(k_s - k_j\right)
\end{split}
\end{equation}
We efficiently optimize our energy function \(\textbf{E}(M)\) using a \emph{conditioned gradient ascent} approach. The general gradient ascent method is a first-order 
optimization procedure that aims at finding local maxima by taking steps proportional to the energy gradient. The conditioner is a scalar factor associated to the 
analytical derivatives that increases (resp. decreases) step-by-step when the gradient sign is constant (resp. fluctuating). The use of the conditioner brings 
three main advantages:  it allows for faster convergence to the final solution, it prevents typical zig-zag-ing while approaching local maxima, and it constraints 
at the same time the analytical derivative size. 

%-------------------------------------------------------------------------
%\subsection{Temporal Smoothing}
%\todo{refer to the video for visual results - cannot display this in simple images.}

%-------------------------------------------------------------------------
\section{Results}
\label{section:results}
We tested our approach on three different datasets: $skirt$, $dance$ and $pop2lock$. Input multi-view video sequences,
as well as camera settings and initial coarse mesh reconstruction were provided by Gall \etal~\cite{Gall2009} and Starck and Hilton~\cite{StarckHilton2007}.
All the sequences are recorded with 8 synchronized and calibrated cameras and number of frame ranging between 250 and 721 (see Table~\ref{table:computationTimeSequences}).
The input provided coarse mesh are obtained utilizing low-quality refining technique based on sparse feature matching, shape-from-silhouette and multi-view 3D
reconstruction, and therefore lack of surface details.

\begin{table}
\begin{center}
\begin{tabular}{|l|c|c|c|c|c|}
\hline
Sequence & Frames & $G_s$ & Iter/s & Frame/min \\
\hline\hline
$skirt$ & 721 & 3053 & 2.01 & 0.8 \\
$dance$ & 573 & 3430 & 1.90 &  0.76\\
$pop2lock$ & 250 & 3880 & 1.67 & 0.66\\
\hline
\end{tabular}
\end{center}
\caption{Computation time for the input sequences. The table shows the amount of frames for each sequence, as well as amount of 3D surface Gaussians $G_s$,
iteration per second and frames per minute. %The latter is the upperbound that is reached when all steps until the maximum number 
%of steps are iterated (set to 150). However, in general the local maxima is reached in less than a hundred iterations.
}
\label{table:computationTimeSequences}
\end{table}

In order to refine the input mesh sequences, we first subdivide the input coarse topology, by inserting additional triangles and vertices, aiming at increasing the scale level of detail.
%We obtained a total of $N=6002$ vertices (for $dance$), $N=10002$ vertices (for $skirt$) and $N=10006$ vertices (for $pop2lock$).
Then we generate a collection of Gaussians on the surface as explained in Sect.~\ref{sect:overview}.
%The skirt and dance sequences contained N=721 
%and N=573 frames, respectively, and they were recorded with K=8 cameras with image resolution of 1004x1004.  The $pop2lock$ sequence 
%contains N=250 frames and it was recorded with K=8 cameras with image resolution of 1920x1080.  The original meshes have a resolution 
%of 1502 vertices (dance), 2502 vertices (skirt) and 2503 vertices ($pop2lock$).
%As a pre-processing step, the original coarse meshes are first subdivided using the Loop scheme. 
%As a result, the input mesh resolutions used in our system were 6002 vertices (dance), 10002 vertices (skirt) 
%and 10006 vertices ($pop2lock$).
%In addition, 
Since for the input sequences most of the fine-scale deformations happen on the clothing, we decided to
focus on the refinement of those areas, generating surface Gaussians only for the correspondent vertices. 
Table~\ref{table:computationTimeSequences} shows the amount of 3D surface Gaussians created for each sequence.
%Once we got our implicit model representation in place, following the steps of our algorithm, 
%we generate a set of 2D image Gaussians for each camera view and frame, and iteratively
%optimize our energy formulation (Sect.~\ref{sect:refinement}).
%The output of the optimization is a set of displacements that we utilize
%to displace the vertices of the input mesh along the correspondent vertex normal.

When rendering the final resulting mesh sequences, we added an extra epsilon to the computed vertex displacements equal
to the standard deviation of the surface Gaussians used. This is needed in order to compensate for the small surface bias (shrink
along the normal during optimization) that is due to the spatial extent of the Gaussians.
%, that appear as a shrink of the mesh volume in the normal direction. %slightly shrink the mesh surface along the normal direction.
%that makes resulting surfaces appear shrink along the normal direction.

\begin{figure}
\begin{center}
\includegraphics[width=0.26\linewidth]{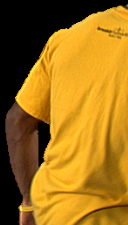}
\includegraphics[width=0.26\linewidth]{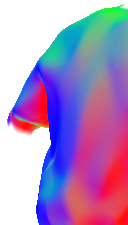}
\includegraphics[width=0.26\linewidth]{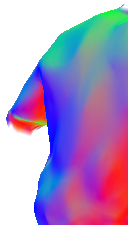}\\
\includegraphics[width=0.26\linewidth]{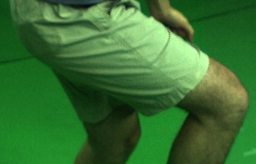}
\includegraphics[width=0.26\linewidth]{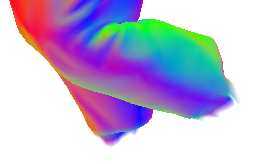}
\includegraphics[width=0.26\linewidth]{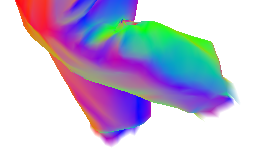}\\
\includegraphics[width=0.26\linewidth]{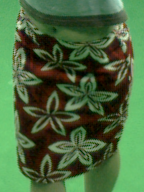}
\includegraphics[width=0.26\linewidth]{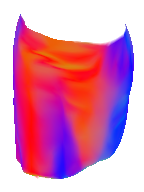}
\includegraphics[width=0.26\linewidth]{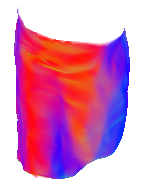}
\end{center}
\caption{Comparison of the results of our refinement capture method against the baseline provided by~\cite{Gall2009,StarckHilton2007} for the $pop2lock$ (top), $dance$ (middle) and $skirt$ (bottom) sequences.  From left to right: input image, color-coded normals of the input mesh and color-coded normals of the rendered output refined mesh.}
\label{figure:results}
\end{figure}

\textbf{Evaluation.}
Our results (Fig.~\ref{figure:teaser}, Fig.~\ref{figure:results} and the accompanying video) show that our 
approach is able to plausibly reconstruct more fine-scale details, \eg the wrinkles and folds in the skirt, and produces closer model alignment to the images than the baseline methods (\cite{Gall2009,StarckHilton2007}).

\begin{figure}
\begin{center}
\includegraphics[width=0.927\linewidth]{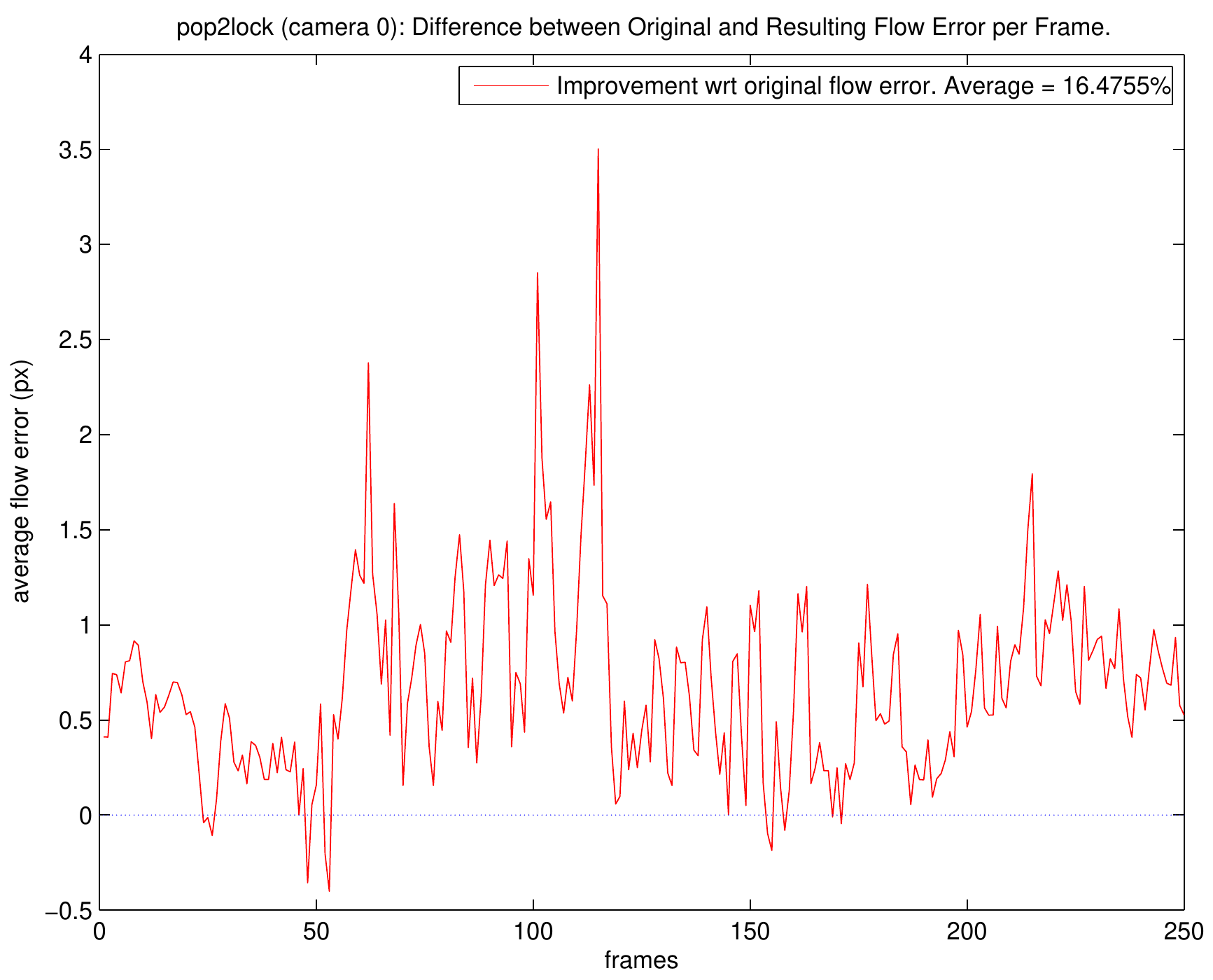}
\includegraphics[width=0.927\linewidth]{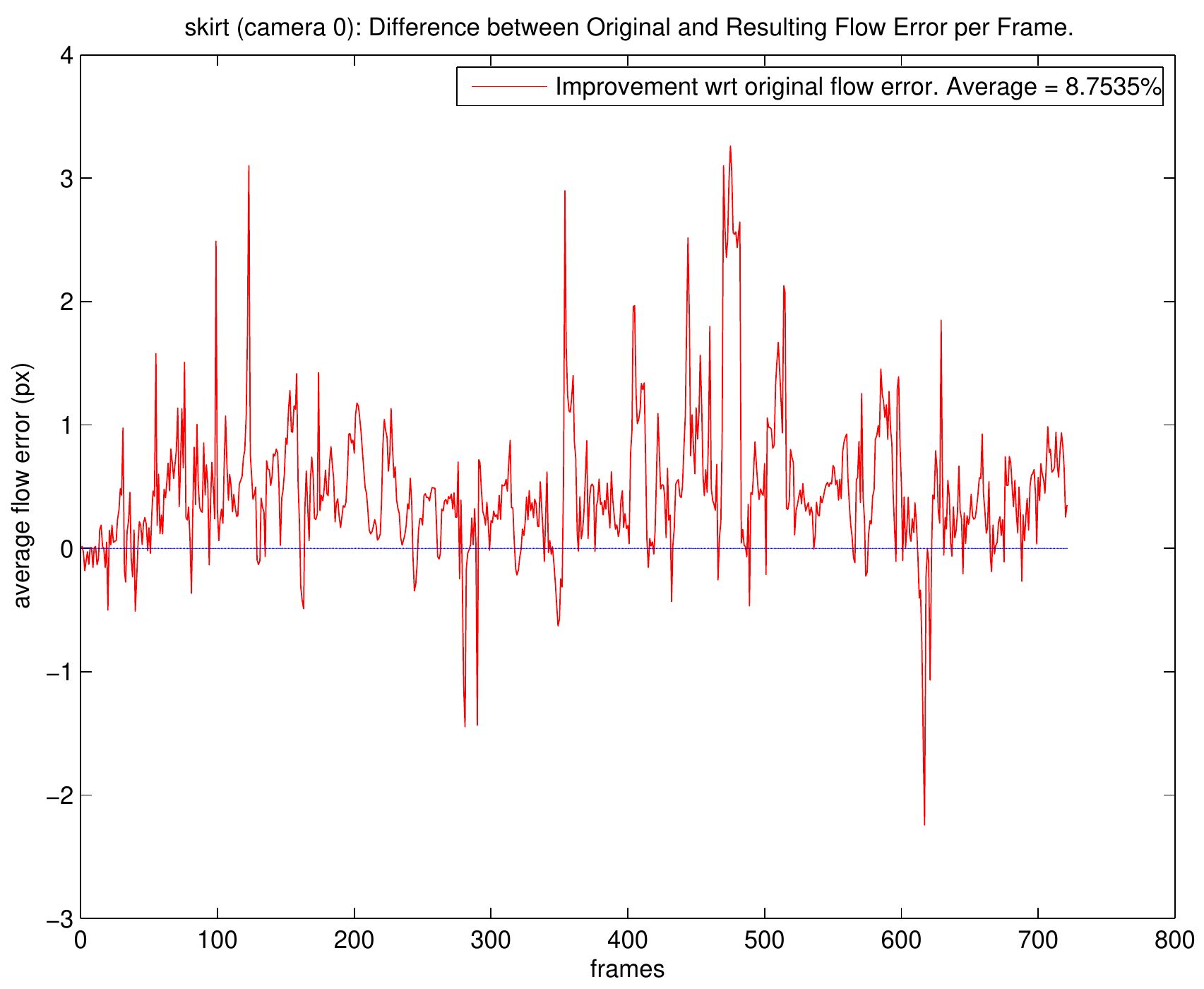}
\includegraphics[width=0.927\linewidth]{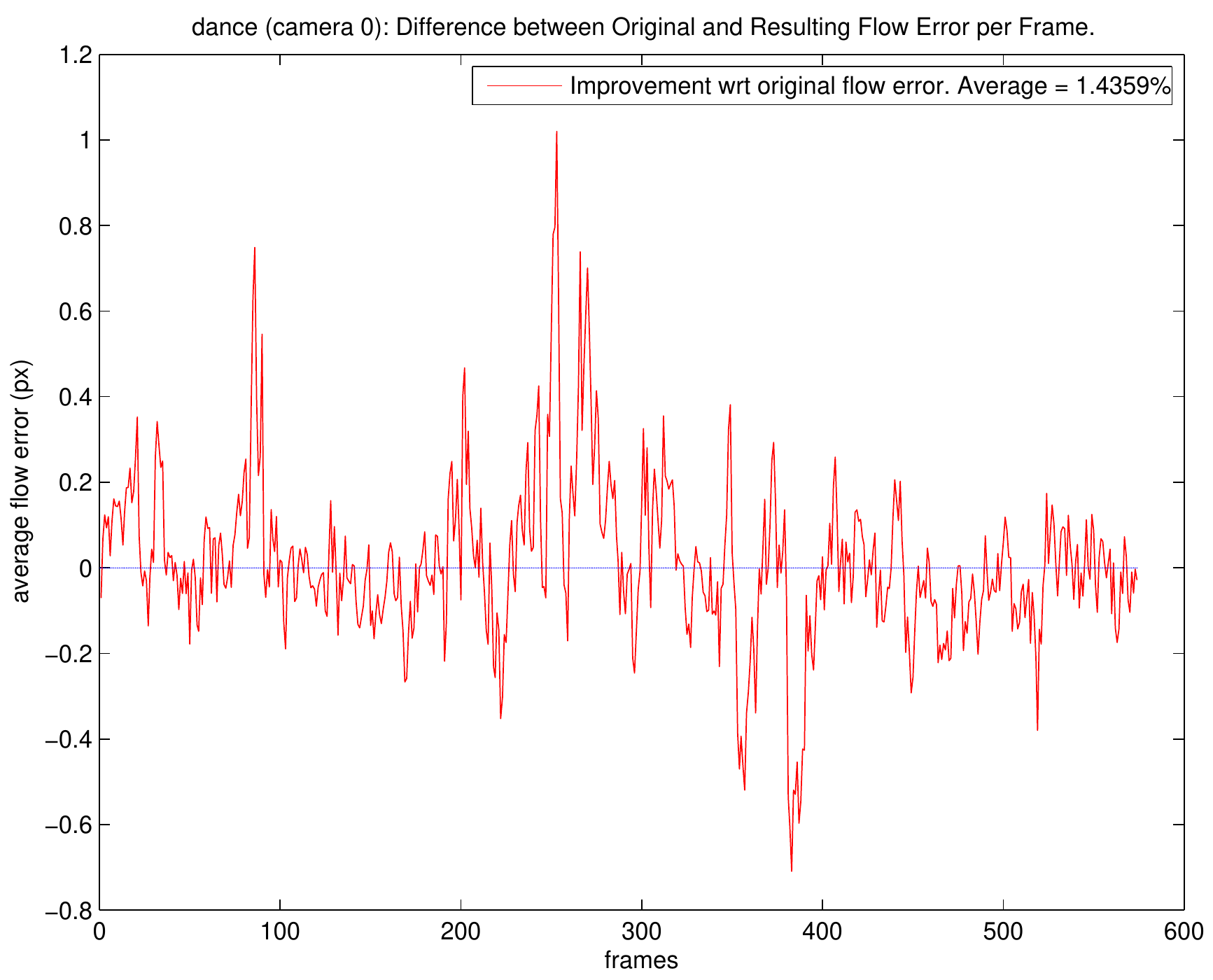}
\end{center}
   %\caption{For the skirt sequence, our refined model (blue curve) presents less mean flow displacement errors when compared to the original input mesh (red curve). 
   %In the figure, the x-axis represents the frame number and y-axis represents the average norm of the optical flow displacement error between 
   %input image and reprojected textured model.}
   \caption{Difference between original and refined sequences in terms of the average optical flow displacement error (in pixels) per frame. We register an average improvement
   equal to $16.4\%$ for the $pop2lock$ sequence (top), $8.7\%$ for the $skirt$ sequence (middle) and $1.4\%$ for the $dance$ sequence (bottom).}
\label{figure:results2}
\end{figure}

%In order to verify the quantitative performance of our approach, we textured the model with the images of the first frame.
In order to verify the quantitative performance of our approach, we textured the model by assigning surface Gaussians colors to the correspondent mesh vertices.
Then, we used optical flow
%~\cite{Bro04a}
to generate displacement flow vectors between the input images 
and the reprojected textured mesh models (original and refined) for all time steps.  Fig.~\ref{figure:results2} plots the average optical flow displacement error difference
between the input and the resulting animation sequences over time for a single camera view. As shown in the graphs, our method decreases the average flow displacement error,
leading to quantitatively more accurate results.

\begin{figure*}
\begin{center}
\includegraphics[width=0.111\linewidth]{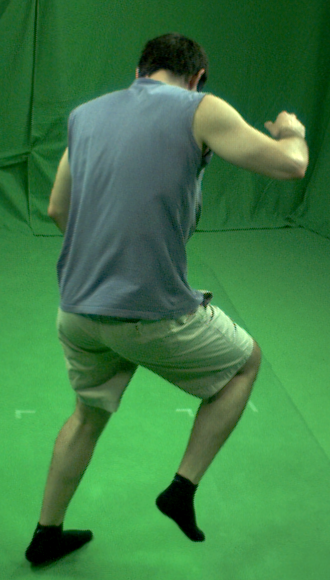}\hspace{10pt}
\includegraphics[width=0.111\linewidth]{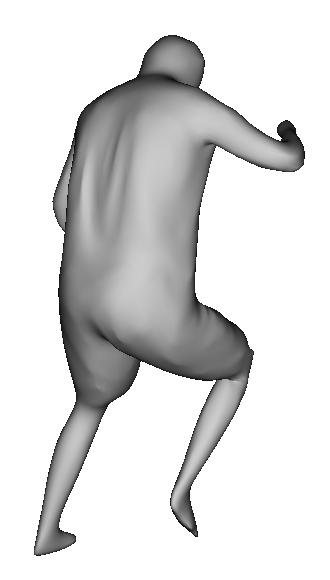}\hspace{10pt}
\includegraphics[width=0.111\linewidth]{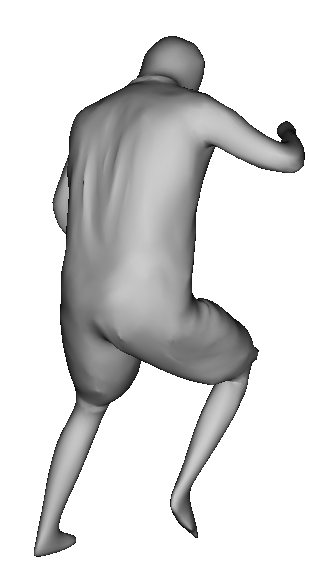}\hspace{10pt}
\includegraphics[width=0.211\linewidth]{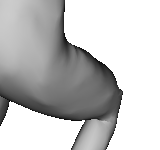}\hspace{10pt}
\includegraphics[width=0.211\linewidth]{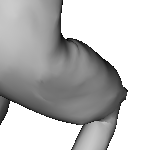}\\
\includegraphics[width=0.111\linewidth]{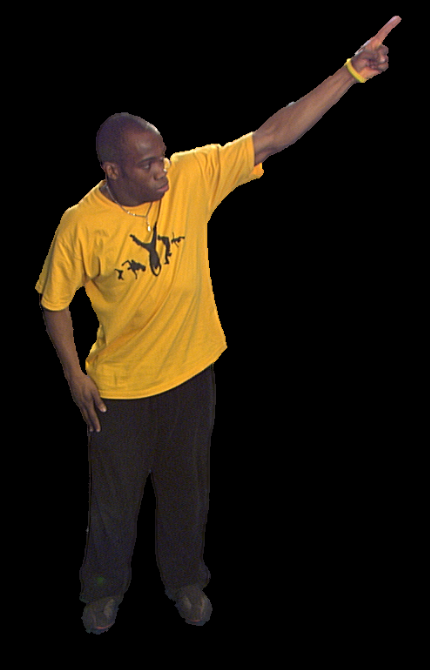}\hspace{10pt}
\includegraphics[width=0.111\linewidth]{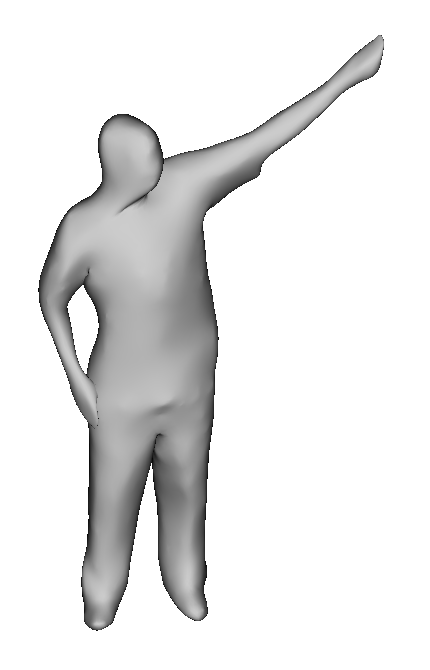}\hspace{10pt}
\includegraphics[width=0.111\linewidth]{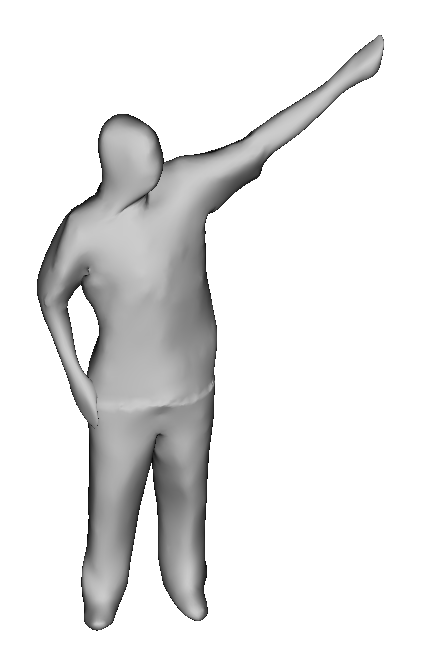}\hspace{10pt}
\includegraphics[width=0.211\linewidth]{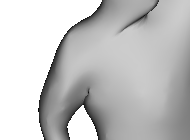}\hspace{10pt}
\includegraphics[width=0.211\linewidth]{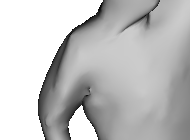}\\
\includegraphics[width=0.111\linewidth]{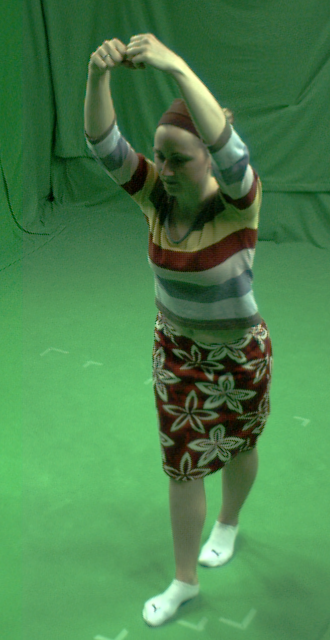}\hspace{10pt}
\includegraphics[width=0.111\linewidth]{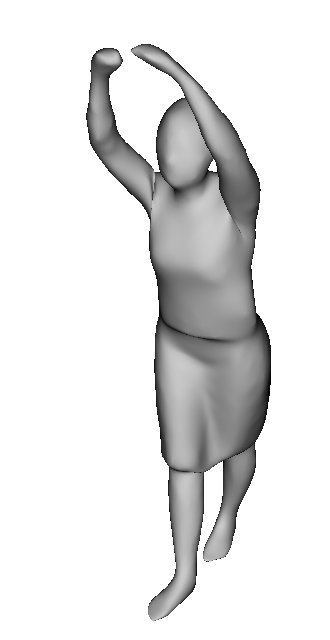}\hspace{10pt}
\includegraphics[width=0.111\linewidth]{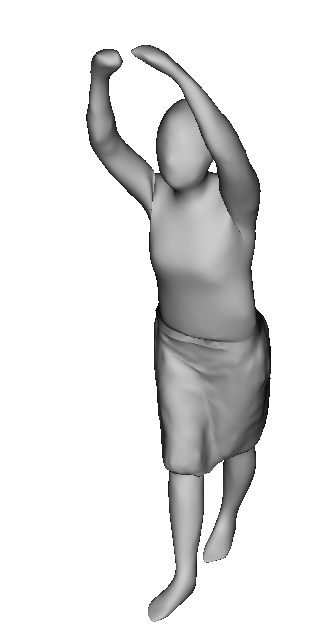}\hspace{10pt}
\includegraphics[width=0.211\linewidth]{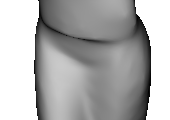}\hspace{10pt}
\includegraphics[width=0.211\linewidth]{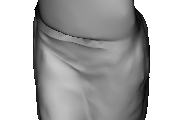}
\end{center}
   \caption{Results of our refinement capture method for the smoothed $dance$ (top), $pop2lock$ (middle) and $skirt$ (bottom) 
   animation sequences.  From left to right: input image, rendered input mesh, rendered output refined mesh, zoom at 
   the rendered input mesh, and zoom at the rendered output refined mesh.}
\label{figure:results1}
\end{figure*}

We created an additional experiment to verify the performance of our refinement framework.  
%For this experiment, we processed the input mesh animations using the method of Desbrun et al.~\cite{Desbrun1999} to first spatially-smooth them, eliminating most of the baked-in surface details.
For this experiment, we first spatially-smooth the input mesh sequence aiming at eliminating most of the baked-in surface details, if any.
The smooth mesh animation is then used as input to our system.  
As we show in Fig.~\ref{figure:results1} and in the accompanying video, our approach is able to plausibly refine the input smooth 
mesh animation, reconstructing fine-scale details in the skirt, t-shirt and shorts. Quantitative evaluation for the smooth input sequence
is provided in an additional document.

%Because we evaluate the energy only for the closer image Gaussians, up to a certain distance threshold always set
%to \(30px\), the maximum number of comparisons is always less than $n_i^M=1200$ for each Gaussian
%on the surface \(\in \{0\dots n_s-1\}\). In the worst case the evaluation of the optimization function, for each converging iteration step, requires
%\(n_i^M n_s n_c\) computations, where \(n_c\) is the number of cameras.
%The importance of our regularization term for the reconstruction of fine details is demonstrated in Fig.~\ref{figure:regularization}. If a larger set of weights is used, the 
%3D surface Gaussians are not able to deform appropriately and details can not be reconstructed properly.  Increasing the value of the weights allows for more 
%deformations of the vertices, but if a larger weight is used, the result is noisy.  For all our experiments, we used the set of weights \(M_d = 2\) and \(w_{reg} = 1\). 
We evaluated the performance of our system on an Intel Xeon Processor E5-1620, Quad-core with Hyperthreading and 16GB of RAM. 
%We spot a sensible improvement of 400\% in speed by only parallelizing the double loop that computes gradients using OpenMP.
%Specifically for about 3000 model blobs we jump from \(0.2\) to \(0.8\) iterations per second. By considering only the closer
%image Gaussian to the model blobs, we further report an improvement of 200\%, moving to \(1.9\) iterations per second.
Table~\ref{table:computationTimeSequences} summarizes the performances we obtained for the three tested sequences. 
We believe we can further reduce the computation time by parallelizing orthogonal steps and implementing our method on GPU.
%We believe we can further reduce the computation time by parallelizing pre-processing operations. Additional computational 
%improvement can be obtained by excluding, when possible, some 3D surface gaussians from the optimization. In fact, 3D 
%surface gaussians that project far from a given camera view or those placed on homogeneously colored surfaces away 
%from the boundaries are likely to produce small errors and gradients that do not contribute sufficiently to the 
%maximization of the energy function. 
%Our approach is subject to a few limitations. We restrict the motion of the surface Gaussians to fall along the 
%normal of the associated vertex. While this is a good assumption, it might lead to collapsing vertices and 
%self-intersections. We also assume that the colors associated to each 3D surface Gaussian is constant and fixed at the first frame.  However, 
%this can be solved by updating the value of the color attribute to each 3D Gaussian over time. The reconstruction quality of our approach 
%depends on the amount of texture and color information in the input video sequence. As a result, 
%sequences with mostly plain colors and few texture regions, like the $pop2lock$ sequence, are not reconstructed with the same level of detail 
%as other sequences, \ie the skirt sequence.  A possible solution here is to include more cues into the optimization framework. 
%Finally, our approach is unable to handle stronger surface deformations as well as topological changes. %We would like to investigate this as a future work.

\textbf{Limitations.}
Our approach is subject to a few limitations.
We assume the input mesh sequence to be sufficiently accurate, such that smaller details can be easily and correctly
captured by simply displacing vertices along their correspondent vertex
normals. In cases where the input reconstructed meshes present misalignments with respect to the images (\eg $pop2lock$) or if it is necessary
to reconstruct stronger deformations, then our method is unable to perform adequately.
In this respect, our refinement should be reformulated allowing more complex displacements, \eg without any normal constraint.
However such weaker prior on vertices motion requires more complex regularization formulation in order to maintain smooth surface, also to handle
unwanted self-intersections and collapsing vertices.
On top of that the increased number of parameters to optimize for (\ie 3 times more, when optimizing for all 3 vertices dimensions, $x$, $y$ and $z$) would spoil
computational efficiency and raise the probability of getting stack in local maxima solutions.
The risk of returning local maxima solutions is still high when employing local solvers (\eg gradient ascent) on non-convex problems as in our case. A possible
solution is to use more advanced solvers, \eg global solvers, when computational efficiency is not a requirement.

Another limitation of our approach is the inability to densely refine plain colored surfaces with few texture (\eg $pop2lock$ and $dance$). 
A solution here is to employ a more complex color model that takes into account \eg illumination and shading effects, at the cost of increased computational expenses.
We would like to investigate these limitations as a future work.
%This kind of motion might considerably restrict the amount of deformations 
%We are unable to model more complex surface deformations
%collapsing vertices and self-intersections
%colors are constant
%texture-less
%local minima
%input mesh sequences
%local optimizer - non-convex energy - run incour 
%We believe we can further reduce the computation time, by parallelizing pre-processing operations as well, such
%as the computation and filtering of the image Gaussian. Additional computational improvement can be obtained by
%excluding, when possible, some model blobs from the energy maximization. In fact, blobs that
%project far from a given camera view or those placed on homogeneously colored surfaces away from the boundaries,
%are likely to produce small error and gradient that do not contribute sufficiently to the maximization of the energy.
%In the case of surface with homogeneous color, the model blobs that give most of the high-frequent information are
%effectively the ones placed on the borders or close to them.
%Despite these limitations, we have presented an efficient framework for performance capture of deforming meshes with 
%fine-scale surface detail using an implicit representation for the deformable mesh and input videos.

%-------------------------------------------------------------------------
\section{Conclusions}
We presented a new effective framework for performance capture of deforming meshes with fine-scale 
time-varying surface detail from multi-view video recordings. 
Our approach captures the fine-scale deformation of the mesh vertices by maximizing photo-consistency on all vertex positions. 
This can be done efficiently by densely optimizing a new model-to-image consistency energy function that uses our proposed implicit representation 
of the deformable mesh using a collection of 3D Gaussians for the surface and a set of 2D Gaussians for the input images. 
Our proposed formulation enables a smooth closed-form energy with implicit occlusion handling and analytic derivatives. We qualitatively
and quantitatively evaluated our refinement strategy on 3 input sequences, showing that we are able to capture and model finer-scale details.

%Our approach can be used to efficiently and densely refine input mesh sequences, obtained from coarse-to-fine performance capture approaches.

%-------------------------------------------------------------------------

{\small
\bibliographystyle{ieee}
\bibliography{egbib,perfcap_bib}
}

\end{document}